\documentclass[arxiv]{meowreport}

\usepackage{algorithm}
\usepackage{algpseudocode}
\usepackage{booktabs}
\usepackage{graphicx}
\usepackage[table]{xcolor}

\definecolor{pos}{RGB}{34,139,34}
\definecolor{neg}{RGB}{200,50,50}
\meowtitle{MeetingToM: Evaluating Multimodal LLMs on \\Theory-of-Mind Reasoning in Multi-Party Meetings}
\meowauthors{Ziyi Wang\affmark{1}\equalcontrib \quad
Yuhang Wu\affmark{1}\equalcontrib \quad 
Dongxu Piao\affmark{1} \quad 
Xingyu Liu\affmark{1} \quad
Tianhui Zhou\affmark{2} \quad
Miao Liu\affmark{1}\corrauth 
}

\meowaffiliation{\affmark{1}College of AI, Tsinghua University\\
\affmark{2}Department of Biostatistics and Bioinformatics, Duke University\\}
\meowdate{July 20, 2026}
\meowlinks{
  \meowlink{Project Page}{https://oliviaziyi.github.io/MeetingToM-Project-Page/}
  \quad
  \meowlink{Code}{https://github.com/oliviaziyi/MeetingToM}
  \quad
  \meowlink{Dataset}{https://huggingface.co/datasets/OliviaWang1101/MeetingToM}
}

\begingroup

\footnotetext[1]{Equal contribution.}
\endgroup

\begin{document}

\makemeowtitle

\begin{meowabstract}
Theory of Mind (ToM), the ability to infer other's beliefs, intentions, and states of knowledge, is central to social interaction, yet remains challenging for current Multimodal Large Language Models (MLLMs), especially in multi-party meetings where cues are distributed across speech and behavior. Existing multimodal ToM benchmarks mainly focus on video-grounded question answering over overt, externally verifiable signals, and provide limited coverage of latent social states and group dynamics. We introduce \textbf{MeetingToM}, a benchmark for complex social behavior reasoning in naturalistic multi-party meetings. MeetingToM targets meeting-specific phenomena such as \textbf{pseudo-consensus}, where apparent agreement masks private dissent under social pressure. The benchmark is hierarchically organized to evaluate ToM at increasing levels of social granularity, including (i) subject-level mental state prediction, (ii) dyadic-level addressee understanding, and (iii) group-level consensus reasoning. We provide a unified evaluation protocol and conduct systematic analyses of representative MLLMs, revealing persistent limitations in integrating non-verbal cues, inferring hidden attitudes, and distinguishing genuine consensus from pseudo-consensus. Our results highlight key challenges and establish MeetingToM as a testbed for advancing meeting-grounded ToM in multimodal models.
\end{meowabstract}


\section{Introduction}

Humans continuously infer and update their own and others’ beliefs, intentions, and states of knowledge during social interactions. This capability, known as Theory of Mind (ToM), is fundamental to social behavior reasoning. Prior work shows that ToM is context-sensitive~\cite{speiger2025evidence}, with goals, plans, and social norms shaping how people reason about others’ belief states. Notably, these contextual factors are especially salient in multi-party meetings, where participants bring distinct goals, partial knowledge, and social constraints. Accordingly, we seek to probe computational Theory of Mind in meeting scenarios. An AI system capable of interpreting complex social states in such a setting would enable practical applications such as collaborative decision support and socially aware assistants.

\begin{figure}[t]
  \centering
  \includegraphics[width=\linewidth]{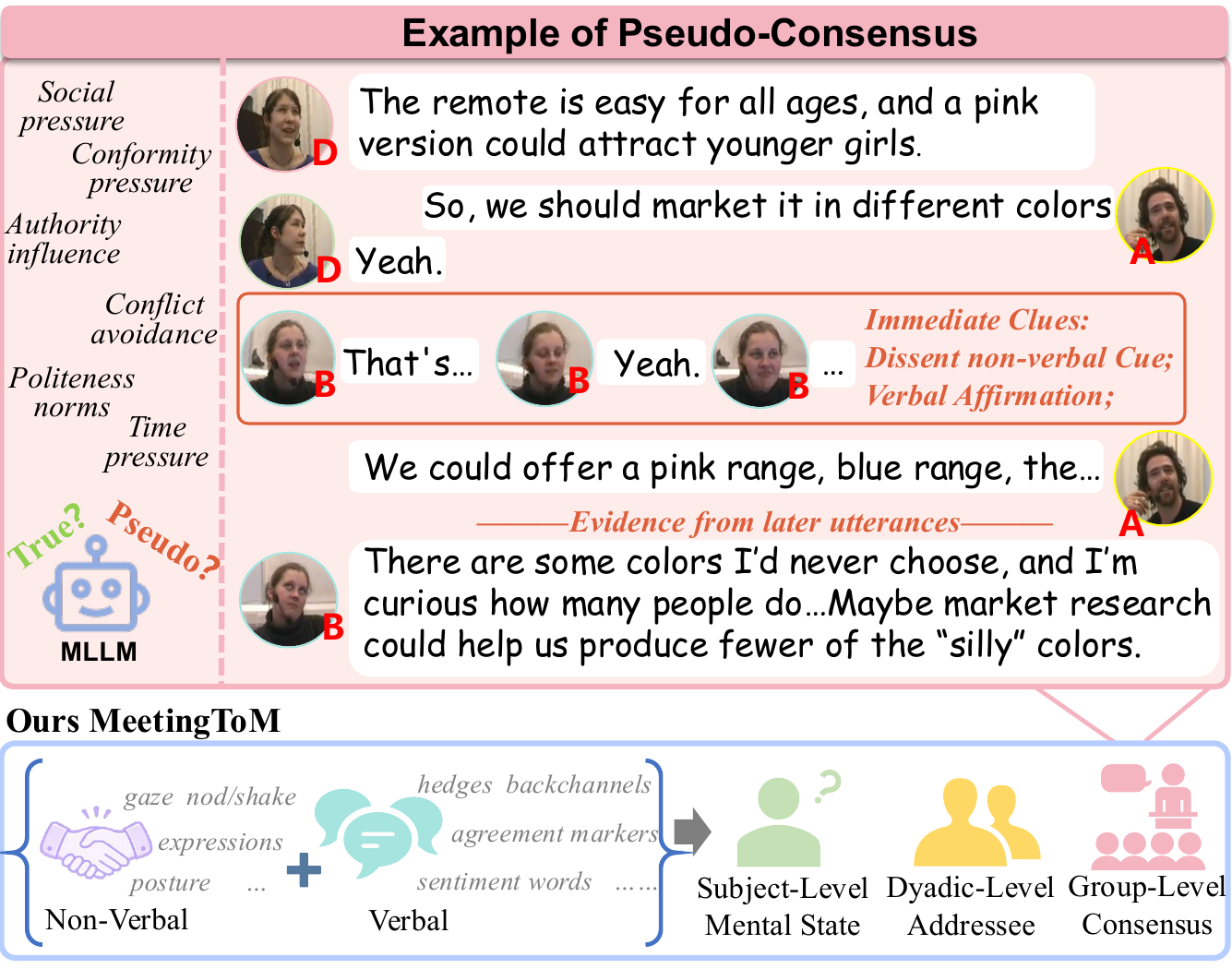}
  \caption{\textit{Hierarchically structured tasks in our MeetingToM benchmark and example of pseudo-consensus.} MeetingToM evaluates meeting-grounded ToM reasoning at three levels: subject-level mental states, dyadic addressee identification and attitude inference, and group-level consensus reasoning. The example shows pseudo-consensus, where verbal agreement masks non-verbal dissent under social pressure.}
  \label{fig:teaser}
\end{figure}

With the advent of Multimodal Large Language Models (MLLMs), there has been growing interest in probing their capabilities for Theory of Mind from a multimodal perspective, motivated by the fact that beliefs and intentions are often conveyed through both verbal and non-verbal cues. Existing multimodal ToM benchmarks~\cite{jin2024mmtomqa,shi2025mumatom} largely emphasize video-grounded question answering, often focusing on overt, externally verifiable signals, rather than systematically testing nuanced, latent social states that unfold in realistic group interactions.

Multi-party meetings, however, present a particularly challenging and consequential setting for multimodal ToM. Beyond understanding what is said, models must piece together evidence from speech and behavior, including turn-taking, gaze, posture, and how people respond to each other~\cite{pentland2008honest}, while considering norms such as politeness and conformity. Crucially, meetings frequently contain surface-level agreement that masks private reservations: participants may verbally endorse proposals while subtly signaling doubt or disengagement, and disagreements may only become apparent through subsequent actions or indirect cues. Such phenomena call for benchmarks to explicitly evaluate whether models can infer hidden attitudes and group-level social dynamics, including cases where consensus is only apparent.

To this end, we introduce \textbf{MeetingToM}, a benchmark for evaluating MLLMs’ capabilities for complex social behavior reasoning in naturalistic multi-party meetings. Consider the example in Fig.~\ref{fig:teaser}. We present a case in which participant B verbally agrees with another participant, yet their body language suggests otherwise, and subsequent utterances reveal underlying concerns. This social phenomenon, where apparent agreement masks private dissent under social pressure, is closely related to groupthink and conformity effects~\cite{janis2008groupthink}. In MeetingToM, we refer to this meeting-specific manifestation as \textbf{pseudo-consensus}. Concretely, our benchmark tasks are hierarchically organized, with increasing demands on ToM capabilities:

\begin{itemize}[leftmargin=*,nosep]
   \item \textbf{Subject-Level Mental State Prediction}: Infers a target participant's mental state from a brief video segment and aligned utterances.
   \item \textbf{Dyadic-Level Addressee Understanding}: Identifies the referent of ``you'' using gaze and body orientation cues, followed by inferring the addressee's attitude.
   \item \textbf{Group-Level Consensus Reasoning}: Determines whether group agreement is genuine, reflects pseudo-consensus, or is absent, and identifies any participants who privately dissent.
\end{itemize}

In our experiments, we first benchmark both proprietary and open-source MLLMs. We then analyze how input modalities, thinking prompts, discourse markers, and contextual priors affect meeting-grounded social-state understanding.

\section{Related Work}
\label{sec:related}

\noindent\textbf{Social Intelligence in Meeting Scenarios}.\ Early research in social signal processing seeks to quantify group dynamics through observable behaviors such as exchange of conversational turns, recognition of roles, and establishment of dominance hierarchies \cite{Vinciarelli2008SocialSignalsSurvey, pentland2008honest}. Recent studies have transitioned toward modeling higher level social constructs like group cohesion via nonverbal cues~\cite{sanchez2012nonverbal, xu2024socialai}. For example, \cite{Vinciarelli2008SocialSignalsSurvey} modeled social signaling through prosodic features, while~\cite{jayagopi2009modeling} utilized visual attention to infer dominance. Recently,~\cite{ouyang2025multispeakerattentionalignmentmultimodal} proposed a multi-speaker attention alignment framework to characterize social targets in group-level rapport. However, these approaches mostly focus on direct behavioral signals, without reasoning inherent social dynamics, such as pseudo-consensus, that mask private disagreement under social pressure \cite{almaatouq2020adaptive, villa-cueva-etal-2025-moments}. Our benchmark bridges this gap by introducing novel tasks that identify both internal cognitive states and complex social dynamics in meeting scenarios.


\begin{figure*}[t]
  \centering
  \includegraphics[width=0.9\linewidth]{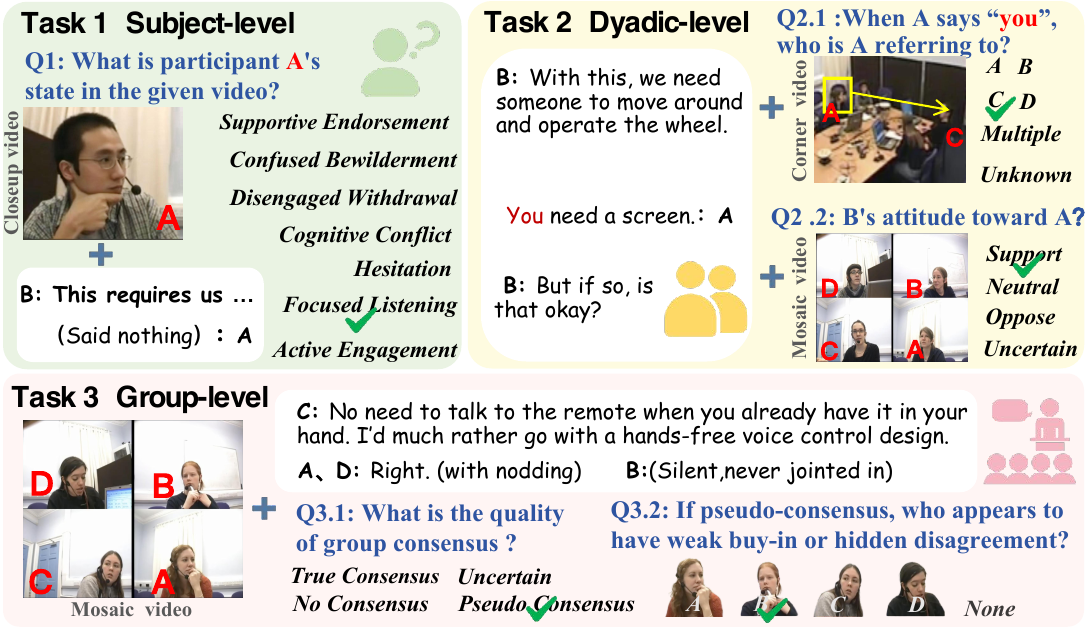}
  \caption{\textit{Overview of the hierarchical tasks in our MeetingToM benchmark.} Subject-level task assesses mental state prediction in meeting scenarios. Dyadic-level Task evaluates addressee identification and interpersonal attitude inference. Group-level tasks probe group consensus reasoning, including consensus quality assessment and the identification of hidden disagreement. These three tasks require progressively richer integration of verbal and non-verbal cues for social reasoning in multi-party interactions.}
  \label{fig:task}
\end{figure*}

\noindent\textbf{Multimodal Social AI Benchmarks}.\ Previous multimodal benchmarks primarily evaluate dyadic social affect recognition \cite{zadeh2018multi, busso2008iemocap, poria2017survey}. Representative examples include Social-IQ \cite{zadeh2019socialiq}, which targets in-the-wild multimodal social question answering, and SIV-Bench \cite{kong2025sivbench}, which focuses on interaction reasoning over social scenes, states, and dynamics. The ToM-oriented resources such as MoMentS \cite{villa-cueva-etal-2025-moments} emphasize narrative-centric mental state inference, while MM-Soc \cite{jin2024mmsoc} targets socially situated online multimodal content. More recent benchmarks introduced in 2025 further push toward evidence-grounded and multi-agent ToM reasoning: Social Genome \cite{mathur-etal-2025-social} evaluates grounded social reasoning with human-annotated reasoning traces over multimodal interactions, and MuMA-ToM \cite{shi2025mumatom} tests embodied multi-agent Theory-of-Mind inference from video and textual contexts. Meeting corpora offer a distinct testbed for multiparty professional collaboration, yet large-scale resources such as MISP 2025~\cite{gao2025multimodal} are largely limited to audio-visual diarization. In contrast, our benchmark probes the in-depth reasoning ability of MLLMs, spanning subject-level internal states, dyadic-level reference understanding, and group-level consensus reasoning.

\noindent\textbf{Theory of Mind in Multiparty Interaction}.\ Theory of Mind (ToM) concerns inferring and tracking latent mental states, including beliefs, intentions, and false beliefs. Early NLP work evaluated ToM mainly via story based QA (e.g., ToM QA Dataset \cite{nematzadeh2018evaluating}), but later analyses showed that common setups can be addressed by exploiting dataset regularities \cite{le2019revisiting}. Recent LLM benchmarks for computational ToM primarily focus on dialogue-based QA, emphasizing higher-order belief recursion \cite{wu2023hitom}, broader skill coverage \cite{chen2024tombench}, and long-form narratives with personality and intention cues \cite{xu2024opentom}.  Broader evaluations with human comparison report notable gaps \cite{kosinski2024evaluating, strachan2024testing} and motivate explicit belief tracking and robustness oriented assessment \cite{sap2022neural, sclar2023minding}. Beyond text modality, ToM is increasingly tested in video grounded QA \cite{jin2024mmtomqa, villa-cueva-etal-2025-moments}, interactive dialogue requiring belief updates \cite{qiu2024minddial, yu2025persuasivetom}, and role play or embodied multi-agent settings \cite{shinoda2025tomato, shi2025mumatom}. Our work shares the motivation of multimodal ToM modeling, with a specific focus on meeting scenarios, where shared goals and task-driven participant roles provide priors that shape how MLLMs model participants’ internal states.

\begin{figure*}[t]
  \centering
  \includegraphics[width=0.9\linewidth]{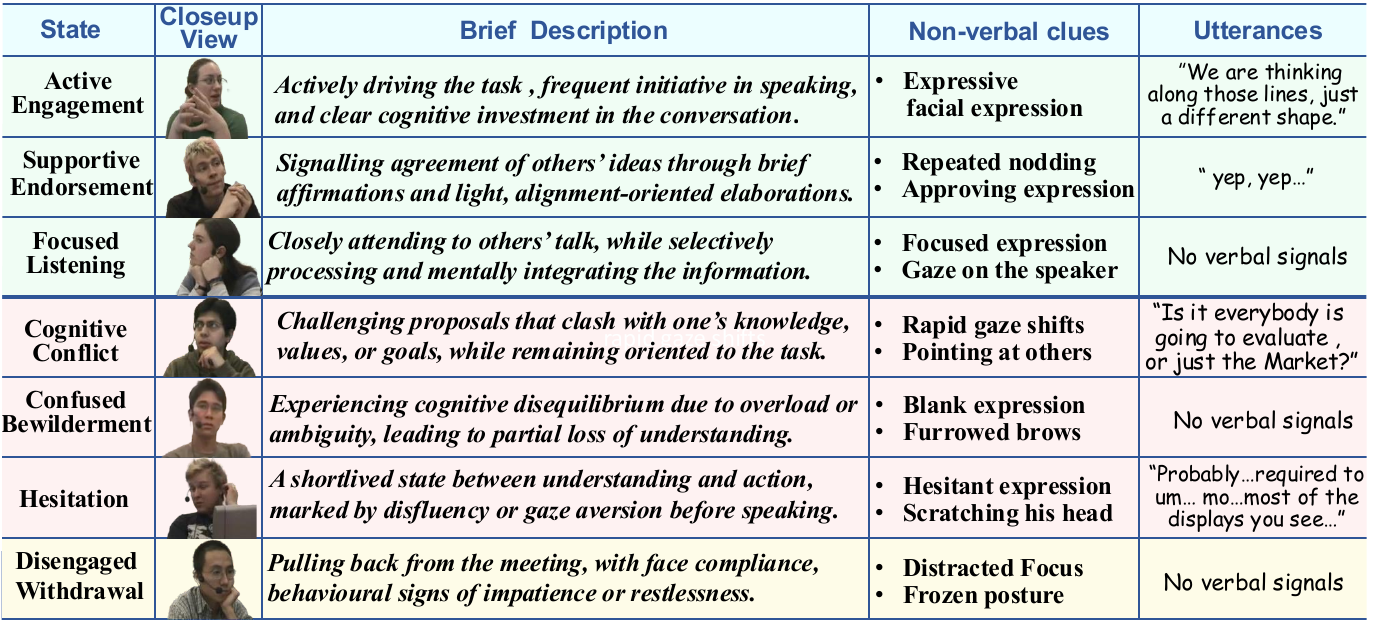}
  \caption{\textit{Illustration of participant cognitive state definitions in meeting scenarios.} For each state, we show a representative close-up view, a brief semantic description, characteristic non-verbal cues (e.g., facial expressions, gaze, and gestures), and example utterances when present. These examples highlight how fine-grained mental states manifest through both verbal and non-verbal cues.}
  \label{fig:state}
\end{figure*}

\section{MeetingToM Benchmark}
\subsection{Overview}
We introduce \textbf{MeetingToM}, a multimodal benchmark designed to evaluate MLLMs’ capability to interpret belief states of participants in naturalistic multi-party meeting scenarios. Our benchmark is motivated by three core observations about real-world meetings: (1) participants continuously attribute their own and others’ mental states; (2) referential expressions require joint modeling of discourse context, speaker intent, and participant roles for referent disambiguation; and (3) apparent group consensus can exert social pressure that masks individual disagreement. These phenomena demand integrated reasoning over visual cues such as facial expressions, gestures, and posture, as well as verbal content such as transcripts. Concretely, MeetingToM comprises three hierarchically structured tasks with progressively increasing Theory of Mind (ToM) reasoning depth:


\begin{itemize}[leftmargin=*,nosep]
    \item Task 1 targets \emph{subject-level} mental state identification with multimodal signals.
    \item Task 2 requires \emph{dyadic-level} understanding about referential intent and interpersonal attitudes.
    \item Task 3 demands \emph{group-level} reasoning about consensus and pseudo-consensus.
\end{itemize}

In the following sections, we first formally define the MeetingToM benchmark tasks and then detail the data sources and annotation process.

\subsection{Task Definition}

\paragraph{Input Notations.} We formalize MeetingToM as a multiple-choice visual question answering benchmark that requires models to select the correct option based on video inputs captured from specific perspectives and accompanying conversational utterances. We denote conversation utterances as $\mathcal{U}$ and the video segment from view $p$ as $\mathcal{V}^p$, where $p \in \{g, A, B, C, D\}$, where $g$ denotes the global view and $A$–$D$ denote the close-up views of the four participants, respectively. We also define a mosaic view $\mathcal{V}^{m}$ by spatially concatenating the synchronized closeup views $\{\mathcal{V}^{A}, \mathcal{V}^{B}, \mathcal{V}^{C}, \mathcal{V}^{D}\}$ into a $2 \times 2$ four-panel video.


\paragraph{Task 1: Subject-level Mental State Prediction.}


This task evaluates whether models can recognize each participant's mental state within a short temporal window, given the corresponding video segment and utterances. Building on prior work in psychology and cognitive science~\cite{apperly2012what}, we define the following meeting mental states as annotation labels:


\begin{itemize}[leftmargin=*,nosep]
    \item \textsc{Active Engagement}: Actively leading or driving the discussion \citep{kahn1990psychological}
    \item \textsc{Supportive Endorsement}: Showing agreement or validation \citep{edmondson1999psychological}
    \item \textsc{Focused Listening}: Attentive but neutral reception \citep{broadbent1958perception}
    \item \textsc{Cognitive Conflict}: Experiencing disagreement or skepticism \citep{dedreu2006conflict}
    \item \textsc{Confused Bewilderment}: Exhibiting confusion or uncertainty \citep{sweller1988cognitive}
    \item \textsc{Hesitation}: Pausing before responding or acting \citep{clark2002using}
    \item \textsc{Disengaged Withdrawal}: Showing low engagement or attention drift \citep{kahn1990psychological}
\end{itemize}

Fig.~\ref{fig:state} presents a visual illustration of these mental states, and detailed definitions are included in Appendix~\ref{app:mental-states}.


Given a short close-up video clip of the target subject $A$, denoted by $\mathcal{V}^{A}$,
and the utterances $\mathcal{U}$, the model predicts $A$'s mental state:
\begin{equation}
\hat{s}_{A}
= \arg\max_{s \in \mathcal{S}}
p\!\left(s \mid \mathcal{V}^{\text{A}}, \mathcal{U}, \mathcal{Q}^{\text{ms}}\right),
\end{equation}
where $\mathcal{S}$ is the set of aforementioned mental states, and $Q^{\text{ms}}$ is the mental state probing query. This task requires models to integrate verbal and nonverbal cues such as speech, facial expressions, and gaze for mental state reasoning.



\paragraph{Task 2: Dyadic-Level Referential Reasoning.}

This task considers two aspects of pair-wise social behavior reasoning based on specified visual views. When a speaker $A$ uses the pronoun “you”, the model performs
\textbf{(i) Addressee Identification} to determine who is being addressed and \textbf{(ii) Attitude Inference} to assess the attitude the addressee holds toward $A$’s statement.

Given the global view $\mathcal{V}^{\text{g}}$ and surrounding utterances $\mathcal{U}$,
and Addressee Identification query$\mathcal{Q}^{\text{addr}}$, this task is defined as
\begin{equation}
\hat{r}
= \arg\max_{r \in \mathcal{R}}
p\!\left(r \mid \mathcal{V}^{\text{g}}, \mathcal{U}, \mathcal{Q}^{\text{addr}}\right),
\label{eq:addr}
\end{equation}
where $\mathcal{R}$ is the addressee label space including:
$A$, $B$, $C$, $D$, \textsc{Multiple}, and \textsc{Unknown}.
Correspondingly, the Attitude Inference task is defined as follows:
\begin{equation}
\hat{m}_{\hat{r}\rightarrow A}
= \arg\max_{m \in \mathcal{M}}
p\!\left(m \mid \mathcal{V}^{\text{m}}, \mathcal{U}, \mathcal{Q}^{\text{att}}, \hat{r}\right).
\label{eq:att}
\end{equation}
where $\mathcal{M}$ is the attitude label space with four classes:
\textsc{Support}, \textsc{Oppose}, \textsc{Neutral}, and \textsc{Uncertain}. $\hat{r}\rightarrow A$ refers to the directional relationship from speaker $A$ to addressee, while $\mathcal{Q}_{\text{att}}$ denotes the attitude inference prompt.

These two tasks are particularly challenging. The Addressee Identification task requires models to reason about gaze direction and body orientation grounded in the specified pronoun and aligned utterances. The Attitude Inference task requires models to infer genuine sentiment from verbal signals and nonverbal cues across stitched views.


\paragraph{Task 3: Group-Level consensus Reasoning.}

This task examines whether MLLMs can identify genuine group agreement or merely surface-level consensus that masks hidden dissent. Specifically, at the group decision-making moment $t^\star$,the model performs
\textbf{(i) Consensus Classification} to assess the quality of agreement and \textbf{(ii) Dissent Identification} to identify the dissenting participant when pseudo-consensus is detected, based on the mosaic view $\mathcal{V}^{\text{m}}$ and the utterances $\mathcal{U}$.
Consensus Classification is defined as
\begin{equation}
\hat{c}
= \arg\max_{c \in \mathcal{C}}
p\!\left(c \mid \mathcal{V}^{\text{m}}, \mathcal{U}, \mathcal{Q}^{\text{cons}}\right),
\end{equation}
where $\mathcal{C}$ is the consensus label space with four classes:
\textsc{TC} (True Consensus), \textsc{PC} (Pseudo Consensus),
\textsc{NC} (No Consensus), and \textsc{U} (Uncertain).



If $\hat{c}=\textsc{PC}$, the model identifies the hidden dissenter with dissent identification query $\mathcal{Q}^{\text{diss}}$:
\begin{equation}
\hat{n}
= \arg\max_{n \in \mathcal{N}}
p\!\left(n \mid \mathcal{V}^{\text{m}}, \mathcal{U}, \mathcal{Q}^{\text{diss}}\right),
\end{equation}
where $\mathcal{N}$ denotes the set of participants, with \textsc{None} indicating no dissenter is presented.

This task probes into an interesting social phenomenon: in professional settings, social pressure often leads individuals to verbally agree while their nonverbal behavior reveals reservation or disagreement. Detecting such pseudo-consensus requires identifying subtle incongruence between verbal affirmation and dissent visual cues such as crossed arms, delayed responses, or gaze aversion.




\begin{figure*}[t]
  \centering
  \includegraphics[width=0.99\linewidth]{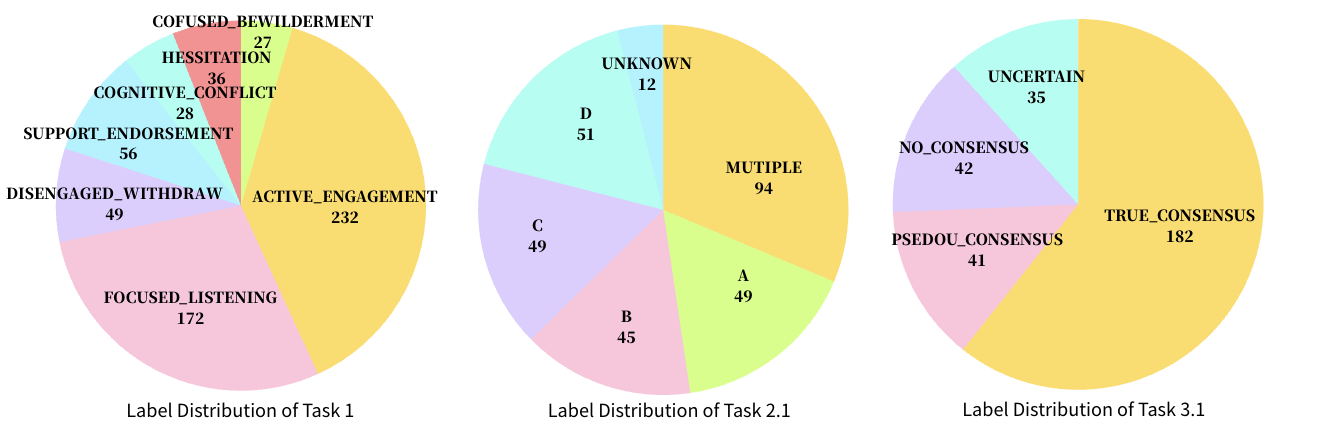}
  \caption{\textit{Per-task label distribution}. We present the label distributions for representative tasks (1, 2.1, and 3.1).}
  \label{fig:statistics}
\end{figure*}

\subsection{Data and Annotation}

\paragraph{Source Data.}
MeetingToM is built on the AMI Meeting Corpus~\cite{carletta2005ami}, which provides synchronized multi-view meeting videos, word-level transcripts, and metadata. We use scenario-based product-design meetings, where four participants discuss proposals and decisions, yielding natural cases of agreement, disagreement, and consensus formation.

\paragraph{Video Clip Extraction.}
We extract task-specific clips around socially relevant events. Task~1 uses 5-second close-up clips of individual participants, while Tasks~2 and~3 use approximately 50-second clips to retain context for referential resolution and consensus reasoning. For each span, we synchronously cut the relevant video views, audio, and transcript using identical timestamps; inputs include global, close-up, or mosaic views depending on the task. Additional details on candidate extraction, clip construction, and multimodal alignment are provided in Appendix~\ref{app:construction_details}.

\paragraph{Annotation Pipeline.}
We use a hybrid pipeline combining task-driven candidate extraction, GPT-assisted question generation, and human answer annotation:
\begin{enumerate}[leftmargin=*,nosep]
    \item \textbf{Candidate Identification}: We identify interaction segments with second-person pronoun usages, and decision-related dialogue patterns.
    \item \textbf{Question Generation}: GPT-5 drafts candidate questions and task-specific options, but does not determine gold answers.
    \item \textbf{Human Annotation}: Two trained annotators independently label each instance after watching the aligned clip and transcript with task-specific guidelines.
    \item \textbf{Adjudication and Filtering}: Disagreements are adjudicated by a third annotator. Final labels are determined by majority voting; cases without majority agreement are discarded, and retained non-unanimous cases are further verified through joint discussion.
\end{enumerate}
Original AMI annotations and metadata are used only for construction and validation. At evaluation time, models receive only the visual input, aligned transcript, and question; gold labels, AMI annotations, and auxiliary metadata are excluded. Further details are provided in Appendix~\ref{app:construction_details}.

\paragraph{Annotation Quality.}
We compute Fleiss' kappa~\citep{fleiss1971measuring} on an agreement subset of 200 instances per task, where all three annotators independently label the same items before consolidation. Task~1 achieves $\kappa=0.7104$; Task~2 achieves $\kappa=0.7319$ for addressee identification and $\kappa=0.5014$ for attitude inference; Task~3 achieves $\kappa=0.5681$ for consensus classification and $\kappa=0.5729$ for dissenter identification. We note that the reported agreement scores provide only a lower-bound estimate of our annotation quality. As described above, final gold labels are obtained through the adjudication, filtering, and joint-verification procedure described above.

  
\paragraph{Benchmark Statistics.}
MeetingToM contains 1,800 clips from 60 meeting sessions, with 600 instances per task family. Figure~\ref{fig:statistics} shows representative label distributions: Task~1 is skewed toward active and attentive states, Task~2.1 is more balanced, and Task~3.1 is dominated by true consensus. We therefore report macro-averaged metrics for imbalanced subtasks.



 \section{Experimental Results}
 
\subsection{Experimental Setup}
  
\paragraph{Evaluated Models.}

We evaluate both proprietary and open-source MLLMs, including GPT-4o~\cite{openai2024gpt4o}, GPT-5~\cite{openai2025gpt5}, Gemini-3 Pro~\cite{google2025gemini3}, and Qwen2.5/3-VL models~\cite{Qwen2.5-VL,qwen3technicalreport}. For video-capable MLLMs, we provide the video clip, aligned transcript, and task prompt. For GPT-5 in our setup, each clip is converted into uniformly sampled frames at 0.5 FPS and paired with the same transcript and prompt. Additional input-formatting, video-ingestion, and prompt-template details are provided in Appendices~\ref{app:video_ingestion} and~\ref{app:prompt_templates}.

\paragraph{Evaluation Metrics.}
We report exact-match accuracy for all tasks, together with a chance baseline under uniform random guessing for each subtask. Task~3.2 is evaluated conditionally on correctly answering Task~3.1, since dissenter identification is only meaningful when pseudo-consensus is detected. Since Tasks~1 and~3.1 exhibit substantial class imbalance, we also include Macro-F1 for these two tasks in Table~\ref{tab:overall_results}.



\subsection{Main Results}

\begin{table}[t]
\centering
\small
\caption{
\textit{Overall performance of prevailing MLLMs on our MeetingToM benchmark.}
Human performance remains substantially higher across tasks, while proprietary models generally outperform the evaluated open-source models. Macro is reported for imbalanced tasks.
}
\renewcommand{\arraystretch}{1.15}
\resizebox{\linewidth}{!}{
\begin{tabular}{l cc c c cc c}
\toprule
Model &
\multicolumn{2}{c}{Task~1} &
Task~2.1 &
Task~2.2 &
\multicolumn{2}{c}{Task~3.1} &
Task~3.2 \\
\cmidrule(lr){2-3}
\cmidrule(lr){6-7}
& Acc. & Macro & Acc. & Acc. & Acc. & Macro & Acc. \\
\midrule
chance-level 
& 14.29 & 14.29 & 16.67 & 25.00 & 25.00 & 25.00 & -- \\
Human
& 86.33 & 75.79 & 86.00 & 83.33 & 80.33 & 74.97 & 79.67 \\
\midrule
\multicolumn{8}{l}{\textit{Proprietary models}} \\
Gemini-3 Pro   
& 59.00 & 30.39 & 74.33 & 51.67 & 40.67 & 22.13 & 88.52 \\
GPT-4o         
& 36.06 & 14.23 & 29.10 & 46.15 & 21.28 & 10.13 & 58.73 \\
GPT-5          
& 55.67 & 28.69 & 54.33 & 51.00 & 43.00 & 28.99 & 90.70 \\
\midrule
\multicolumn{8}{l}{\textit{Open-source models}} \\
Qwen2.5-vl-32B 
& 55.50 & 26.57 & 37.00 & 47.67 & 18.00 & 9.90 & 51.85 \\
Qwen3-vl-8B    
& 42.17 & 23.74 & 37.67 & 47.67 & 23.33 & 11.19 & 65.71 \\
Qwen3-vl-32B   
& 38.00 & 19.89 & 64.67 & 48.67 & 27.33 & 12.91 & 74.39 \\
\bottomrule
\end{tabular}
}
\label{tab:overall_results}
\end{table}

\begin{table}[t]
\centering
\small
\caption{\textit{Ablation of input modalities using Gemini-3 Pro.} Visual cues from video and verbal cues from utterances contribute differently across tasks, suggesting that reliable multimodal fusion remains challenging.}
\setlength{\tabcolsep}{3pt}
\renewcommand{\arraystretch}{1.15}
\begin{tabular}{l c c c c c}
\toprule
Methods & Task1 & Task2.1 & Task2.2 & Task3.1 & Task3.2 \\
\midrule
Video   & 48.17 & 57.33 & 44.67 & 38.33 & 82.61 \\
Utter.    & 61.33 & 72.67 & 55.00 & 55.67 & 97.60 \\
Video \& Utter.  & 59.00 & 74.33 & 51.67 & 40.67 & 88.52 \\ 
\bottomrule
\end{tabular}
\label{tab:modality}
\end{table}


\begin{table*}[t]
\centering
\scriptsize
\caption{
Performance comparison across prompting strategies. 
Macro is reported for imbalanced tasks; arrows show absolute changes from No Prompt 
($\uparrow$ improvement, $\downarrow$ degradation). 
Task-CoT denotes task-aligned CoT.
}
\setlength{\tabcolsep}{3pt}
\renewcommand{\arraystretch}{1.10}

\resizebox{\textwidth}{!}{%
\begin{tabular}{l c cc c c cc c}
\toprule
Model &
Prompt &
\multicolumn{2}{c}{Task~1} &
Task~2.1 &
Task~2.2 &
\multicolumn{2}{c}{Task~3.1} &
Task~3.2 \\
\cmidrule(lr){3-4}
\cmidrule(lr){7-8}
& & Acc. & Macro & Acc. & Acc. & Acc. & Macro & Acc. \\
\midrule

\multirow{3}{*}{Gemini-3 Pro}
  & No Prompt
  & 59.00 & 30.39
  & 74.33
  & 51.67
  & 40.67 & 22.13
  & 88.52 \\
  & Task-CoT
  & 59.67 \textcolor{pos}{$\uparrow$0.67} 
  & 27.48 \textcolor{neg}{$\downarrow$2.91}
  & 70.67 \textcolor{neg}{$\downarrow$3.66}
  & 56.00 \textcolor{pos}{$\uparrow$4.33}
  & 58.67 \textcolor{pos}{$\uparrow$18.00} 
  & 23.40 \textcolor{pos}{$\uparrow$1.27}
  & 97.73 \textcolor{pos}{$\uparrow$9.21} \\
  & ECoT
  & 61.33 \textcolor{pos}{$\uparrow$2.33} 
  & 25.16 \textcolor{neg}{$\downarrow$5.23}
  & 74.00 \textcolor{neg}{$\downarrow$0.33}
  & 50.67 \textcolor{neg}{$\downarrow$1.00}
  & 55.00 \textcolor{pos}{$\uparrow$14.33} 
  & 23.27 \textcolor{pos}{$\uparrow$1.14}
  & 97.58 \textcolor{pos}{$\uparrow$9.06} \\
\midrule

\multirow{4}{*}{Qwen3-VL-32B}
  & No Prompt
  & 38.00 & 19.89
  & 64.67
  & 48.67
  & 27.33 & 12.91
  & 74.39 \\
  & Naive-CoT
  & 42.17 \textcolor{pos}{$\uparrow$4.17}
  & 25.30 \textcolor{pos}{$\uparrow$5.41}
  & 66.00 \textcolor{pos}{$\uparrow$1.33}
  & 51.00 \textcolor{pos}{$\uparrow$2.33}
  & 32.78 \textcolor{pos}{$\uparrow$5.45}
  & 15.82 \textcolor{pos}{$\uparrow$2.91}
  & 79.59 \textcolor{pos}{$\uparrow$5.20} \\
  & Task-CoT
  & 53.33 \textcolor{pos}{$\uparrow$15.33} 
  & 27.47 \textcolor{pos}{$\uparrow$7.58}
  & 60.00 \textcolor{neg}{$\downarrow$4.67}
  & 49.67 \textcolor{pos}{$\uparrow$1.00}
  & 21.40 \textcolor{neg}{$\downarrow$5.93} 
  & 9.79 \textcolor{neg}{$\downarrow$3.12}
  & 59.38 \textcolor{neg}{$\downarrow$15.01} \\
  & ECoT
  & 54.33 \textcolor{pos}{$\uparrow$16.33} 
  & 27.17 \textcolor{pos}{$\uparrow$7.28}
  & 57.33 \textcolor{neg}{$\downarrow$7.34}
  & 46.67 \textcolor{neg}{$\downarrow$2.00}
  & 37.33 \textcolor{pos}{$\uparrow$10.00} 
  & 18.18 \textcolor{pos}{$\uparrow$5.27}
  & 83.04 \textcolor{pos}{$\uparrow$8.65} \\
\bottomrule
\end{tabular}%
}

\label{tab:prompting}
\end{table*}






\paragraph{Overall Performance.} The performance of both proprietary and open-source models is reported in Table~\ref{tab:overall_results}. For reference, human performance remains substantially higher across all tasks, indicating that the benchmark is well-defined yet challenging for current MLLMs. Among proprietary models, Gemini-3 Pro performs best on Task~1 and the two dyadic subtasks, while GPT-5 achieves the strongest group-level results on Task~3.1 and Task~3.2. GPT-5 also substantially outperforms GPT-4o, especially on Task~1 and the group-level tasks. Open-source models remain competitive on selected subtasks, such as Qwen3-VL-32B on Task~2.1, but generally lag behind the strongest proprietary models across the full benchmark.

However, the leaderboard alone does not tell the full story. The additional Macro scores reveal clear class-imbalance effects: on Task~1, Gemini-3 Pro drops from 59.00 Acc. to 30.39 Macro, while GPT-5 drops from 55.67 to 28.69. This gap suggests that aggregate accuracy overstates model reliability by rewarding performance on frequent or visually salient states while masking failures on minority or context-sensitive states. A similar pattern appears in Task~3.1, where Gemini-3 Pro drops from 40.67 Acc. to 22.13 Macro and GPT-5 drops from 43.00 to 28.99. These gaps indicate uneven performance across consensus states, rather than uniformly reliable group-level reasoning.


\paragraph{Cross-Task Discussion.} Consistent with the hierarchical design of MeetingToM, performance decreases as reasoning becomes more relational and collective. Subject-level mental-state recognition is comparatively easier, while dyadic and group-level tasks require aligning utterances, addressees, non-verbal behavior, and interaction context. For group-level reasoning, Task~3.1 is the main bottleneck; Task~3.2 is evaluated only when the consensus state is correctly identified.

\begin{figure*}[t]
  \centering
  \includegraphics[width=\linewidth]{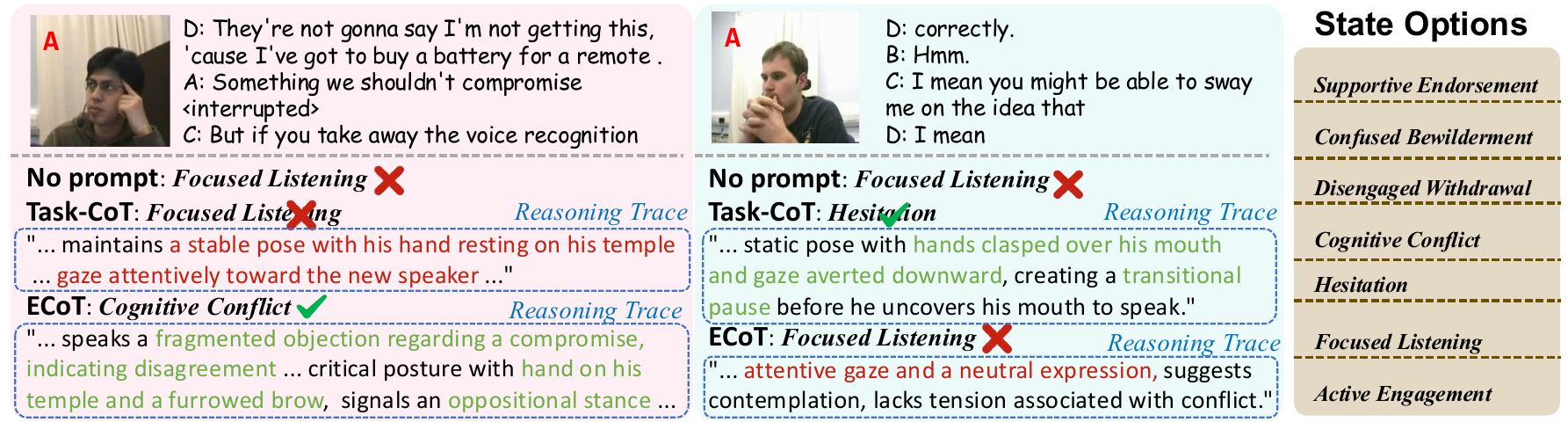}
\caption{\textit{Qualitative examples of prompting effects on subject-level mental state prediction.} The reasoning traces illustrate how ECoT and Task-CoT redirect attention to different social cues.}
  \label{fig:vis}
\end{figure*}

\paragraph{Ablation on Input Modalities.}
We conduct input-modality ablations with Gemini-3 Pro and report the results in Table~\ref{tab:modality}. Utterance-only input performs best on Tasks~1, 2.2, 3.1, and 3.2, suggesting that meeting-grounded social reasoning relies strongly on verbal content and conversational structure. This is especially important for attitude inference and consensus reasoning, where models must track what is said, whether a stance is explicit or implicit, and how participants converge or diverge. However, verbal input alone does not dominate all subtasks. For Task~2.1, combining video and utterances performs best, indicating that addressee identification benefits from aligning linguistic cues with gaze, body orientation, and visual engagement. The joint setting also does not consistently outperform utterance-only input on higher-level tasks, suggesting that current MLLMs may still struggle to fuse visual and verbal evidence when cues are ambiguous. Overall, the modality results suggest that the challenge is not simply whether evidence is available, but whether models can combine it in a task-appropriate way.


\paragraph{Discourse Marker Shortcut Analysis.}
To test whether models rely on shallow transcript shortcuts, we remove hesitation and backchannel markers such as ``hmm'' and ``emm'' while keeping the visual input and task prompt unchanged. As shown in Table~\ref{tab:no_emm_hmm}, removing these markers improves Task~1 but reduces performance on dyadic and group-level tasks. This suggests that such markers may sometimes distract local mental-state recognition, while still carrying useful interactional information for addressee, attitude, and consensus reasoning.


\paragraph{Benefits from Thinking Prompts.}
Thinking prompts can improve some VQA tasks. We test whether they help MeetingToM, where models reason over verbal, non-verbal, and interactional cues. We compare Chain-of-Thought (CoT)~\cite{wei2022chainofthought} and Emotional Chain-of-Thought (ECoT)~\cite{li2024enhancing}. For CoT, we report two variants: Naive-CoT, which uses generic step-by-step reasoning, and Task-aligned CoT (Task-CoT), which adds task-specific guidance for organizing subject-, dyadic-, and group-level social evidence. Prompt templates are provided in Appendix~\ref{app:prompt_templates}.

As shown in Table~\ref{tab:prompting}, prompting effects are task- and model-dependent rather than uniformly beneficial. On Qwen3-VL-32B, Naive-CoT improves all reported subtasks, while ECoT also improves Task~1 and group-level reasoning but reduces both dyadic subtasks. Task-CoT shows a different trade-off: it substantially improves Task~1, but reduces Task~2.1 and group-level performance. Gemini-3 Pro exhibits another pattern, with Task-CoT and ECoT mainly benefiting group-level tasks while producing mixed effects on Task~1 and Task~2. Additional open-source results in Appendix~\ref{app:open_prompting} show that this lack of uniform gains is not specific to Qwen3-VL-32B. These results suggest that thinking prompts do not simply add reasoning depth; they reshape how models allocate attention across social evidence.

Fig.~\ref{fig:vis} illustrates this mechanism qualitatively. ECoT can direct the model toward expression-related cues and improve some subject-level predictions, but it may also over-emphasize facial affect while missing broader interactional evidence. Overall, prompting helps selectively, but does not solve the harder problem of calibrating verbal, visual, and interactional cues under social ambiguity.





\begin{table}[t]
\centering
\small
\caption{Contextual-prior ablation results on Qwen3-VL-32B. Task~1 is not evaluated in this setting ($n{=}0$).}
\setlength{\tabcolsep}{3pt}
\renewcommand{\arraystretch}{1.15}

\begin{tabular}{l c c c c}
\toprule
Methods & Task2.1 & Task2.2 & Task3.1 & Task3.2 \\
\midrule

Qwen3-VL-32B     & 64.67 & 48.67 & 27.33 & 74.39 \\
Base Model w. DA      & 37.94 & 45.74 & 24.08 & 73.61 \\
Base Model w. phase   & 61.87 & 49.83 & 20.74 & 62.90 \\
Base Model w. role    & 51.67 & 51.33 & 24.33 & 67.12 \\

\bottomrule
\end{tabular}

\label{tab:prior_ablation_qwen3vl32b}
\end{table}

\begin{table}[t]
\centering
\scriptsize
\caption{
Discourse-marker ablation on Qwen3-VL-32B.
}
\renewcommand{\arraystretch}{1.15}
\setlength{\tabcolsep}{3pt}
\begin{tabular}{lccccc}
\toprule
Setting & Task~1 & Task~2.1 & Task~2.2 & Task~3.1 & Task~3.2 \\
\midrule
Baseline 
& 38.00 & 64.67 & 48.67 & 27.33 & 74.39 \\
w/o discourse markers 
& 52.50 & 60.20 & 46.49 & 21.48 & 62.50 \\
\bottomrule
\end{tabular}
\label{tab:no_emm_hmm}
\end{table}

\paragraph{Ablations on Contextual Priors.}
We further test whether structured priors improve Qwen3-VL-32B by adding participant role, meeting phase, or dialogue-act (DA) cues. Role, phase, and DA respectively indicate each participant's functional position, the current meeting stage, and each utterance's communicative function, all taken from the original human annotations in the AMI Meeting Corpus. As shown in Table~\ref{tab:prior_ablation_qwen3vl32b}, these priors do not yield consistent gains: DA sharply hurts addressee reasoning and group-level performance, while phase and role provide only isolated improvements. This suggests that structured priors are not automatically useful simply because they are semantically relevant. In meeting-grounded ToM, contextual labels become helpful only when calibrated against local verbal and non-verbal evidence; otherwise, they may be redundant with the video-transcript input or bias the model toward stereotyped expectations. These results point to a deeper bottleneck: current MLLMs struggle not only with missing context, but also with deciding when contextual information should constrain, revise, or yield to situated multimodal evidence.



\section{Conclusion}


We present MeetingToM, a multimodal benchmark for evaluating theory of mind capabilities in naturalistic meeting scenarios. Our benchmark comprises three task families: mental state recognition from short video segments, dyadic social reasoning about referential intent and interpersonal attitudes, and group-level consensus assessment. Experiments with state-of-the-art MLLMs reveal substantial room for improvement, particularly in detecting subtle nonverbal cues and resolving conflicts between verbal and visual signals. We investigate the benefits of chain-of-thought–style prompting and analyze the pitfall of injecting contextual priors. We hope MeetingToM will catalyze progress toward machines that can genuinely understand the social dynamics of human interaction.

\section{Limitations}

First, MeetingToM currently draws from English-language meetings, and thus does not measure MLLMs' performance in multilingual settings. We plan to expand the sources to multilingual meetings in future versions. Second, constrained by current MLLMs' video context limits and the difficulty of annotating long-form interactions, we evaluate models on 5s / 50s key meeting segments rather than full-length meetings (e.g., 30 minutes). MeetingToM also focuses on structured meeting scenarios, which improves controllability and annotation consistency but may limit generalization to informal, online, or hybrid meeting settings. Extending the benchmark to longer meeting-level interactions and platform-mediated meeting contexts remains an important direction for future work; we discuss scope, temporal context, and modern meeting settings in Appendix~\ref{app:scope_limitations}. Third, while we carefully annotate the data in a systematic and consistent manner, we cannot re-contact the original meeting participants to obtain self-reported mental states, so some labels may still differ from the participants' true private beliefs or intentions at the time. We will explore improved data collection protocols to better approximate participant-grounded signals in future work.



\section{Ethical considerations}
MeetingToM is built solely from the publicly available AMI Meeting Corpus, following its documented access and distribution procedures. We do not collect new recordings and do not attempt to re-identify individuals beyond the anonymization already provided by AMI. Our annotations focus on short-lived, episode-bound mental states that are grounded in observable verbal and non-verbal cues within each segment. We explicitly avoid labeling stable personal traits or sensitive attributes. Since private mental states are not directly observable and we cannot obtain participant self-reports, the labels should be interpreted as systematic third-person inferences from available evidence rather than definitive ground truth about participants’ internal experiences. We provide an extended discussion of potential misuse, bias, and deployment boundaries in Appendix~\ref{app:ethics}.

\section{Acknowledgments}
This work was supported in part by the
Zhiyuan Scholar Program of the Beijing Municipal Science and Technology Commission
(Z251100008125045), NSFC Grants, and a research grant from the ByteDance Seed Team.





\bibliographystyle{plainnat}
\bibliography{custom}


\clearpage 
\appendix

This is the supplementary material for the paper "MeetingToM: Evaluating Multimodal LLMs on
Theory-of-Mind Reasoning in Multi-Party Meetings".

\section{Conceptual Background for Meeting Mental States}
\label{app:mental-states}

This section summarizes the conceptual background for the seven meeting-specific mental states used in MeetingToM. The taxonomy is designed to jointly capture affective valence, task engagement, and typical interactional behavior, drawing on work in engagement, cognitive load, affect, and meeting science. Table~\ref{tab:mental-states-compact} provides a compact overview.

\paragraph{Active Engagement.}
Active Engagement is a dynamic cognitive and affective state in which participants invest substantial attentional and emotional resources in the collaborative process, experiencing the meeting as meaningful and worth their effort. It aligns with psychological engagement \citep{kahn1990psychological}, self-determination theory on intrinsic motivation and goal congruence \citep{deci2000whatwhy}, and work on psychological safety and collective efficacy in teams \citep{edmondson1999psychological,bandura2000collective}. In meetings, Active Engagement typically appears as sustained focus, initiative taking, elaborated contributions, and cooperative problem solving, consistent with human performance models and meeting effectiveness research \citep{wickens2021engineering,allen2023keyfeatures,allen2015cambridge}.

\paragraph{Focused Listening.}
Focused Listening is a cognitively active but externally restrained state in which participants selectively attend to others’ speech, integrate new information, and anticipate intentions without yet taking the floor. It is grounded in selective-attention and cognitive-load accounts of information processing \citep{broadbent1958perception,sweller1988cognitive}, as well as work on mentalizing and perspective taking in social cognition \citep{frith2006neural}. Subtle feedback behaviors such as nodding, gaze alignment, or brief backchannels signal this form of engaged reception in meetings and help maintain task alignment and conversational efficiency \citep{tennant2023activelistening,allen2023keyfeatures}.

\paragraph{Supportive Endorsement.}
Supportive Endorsement denotes a positively valenced socio-emotional state of agreement, validation, and affiliative reinforcement toward others’ ideas. It arises under conditions of trust, psychological safety, and perceived goal alignment \citep{edmondson1999psychological,kahn1990psychological}, and is linked to the broadening effects of positive affect on cooperation \citep{WeissCropanzano1996}. Cognitively, it reflects social appraisal focused on consistency and feasibility rather than threat \citep{scherer2001appraisal}; behaviorally, it appears as explicit agreement, praise, and light co-construction that strengthen cohesion and accelerate consensus in meetings \citep{allen2015cambridge}.

\paragraph{Cognitive Conflict.}
Cognitive Conflict is a negatively valenced yet high-engagement state in which participants actively challenge ideas that contradict their knowledge, values, or goals. Organizational research distinguishes such task conflict from relationship conflict and shows that, when managed well, it can improve decision quality and innovation \citep{thomas1992conflict,dedreu2006conflict}. From a cognitive perspective, conflict monitoring and error detection processes are activated \citep{botvinick2001conflict}, while appraisal processes evaluate threats to goals and self \citep{lazarus1991emotion}. In meetings, Cognitive Conflict manifests as counterarguments, requests for justification, and attempts to correct inconsistencies; under psychologically safe climates it can be productive, but it can also escalate into interpersonal strain if mishandled \citep{edmondson1999psychological,allen2023keyfeatures}.

\paragraph{Confused Bewilderment.}
Confused Bewilderment describes a state of cognitive disequilibrium arising when informational demands exceed prior knowledge or when instructions are ambiguous. Cognitive load theory characterizes this overload as a mismatch between task complexity and cognitive resources \citep{sweller1988cognitive}, while affective-events theory explains the accompanying uncertainty and mild negative affect at work \citep{WeissCropanzano1996}. Studies of learning show that confusion can be adaptive, triggering metacognitive monitoring and help-seeking when properly scaffolded \citep{dmello2012dynamics}. In meetings, this state often appears as furrowed brows, delays, or clarification questions, reflecting attentional bottlenecks in processing complex or poorly structured information \citep{broadbent1958perception,allen2023keyfeatures}. Unlike hesitation, confusion primarily reflects difficulty in understanding the task content rather than uncertainty about expressing one’s view.

\paragraph{Hesitation.}
Hesitation is a short-lived cognitive and emotional bridge between uncertainty and overt action, characterized by speech disfluencies, timing delays, and visible self-monitoring before speaking. Psycholinguistic work shows that fillers such as \emph{uh} and \emph{um} signal difficulties in planning and coordination between thought and speech \citep{clark2002using,foxtree1997um}. Appraisal theories interpret hesitation as reflecting real-time checks of adequacy and social-evaluative risk \citep{scherer2001appraisal,lazarus1991emotion}. In meetings, hesitation often precedes key turns in decision making and is observable in micro-pauses, hedging, and gaze aversion, which interact with temporal interaction patterns over the course of a discussion \citep{lehmann2018temporal}. Unlike confusion, hesitation can arise even when the content is understood, reflecting uncertainty about the social consequences of speaking up.

\paragraph{Disengaged Withdrawal.}
Disengaged Withdrawal is a psychologically detached, negatively valenced state in which participants reduce cognitive and emotional investment in the meeting. It builds on engagement and disengagement accounts \citep{kahn1990psychological} and burnout research on chronic strain and cynicism \citep{maslach1997truth}. Cognitively, attention drifts from task content toward thoughts about time and external concerns; affectively, boredom and frustration dominate and are located in low-arousal negative regions of the affective circumplex \citep{russell1980circumplex}. In meeting science, withdrawal is visible as silence, multitasking, or minimal eye contact and is associated with low participation energy and ineffective meeting structures \citep{allen2023keyfeatures,allen2015cambridge,rogelberg2019surprising,WeissCropanzano1996}.

\appendix

\section{Benchmark Construction and Scope}
\label{app:construction_details}

\subsection{Candidate Mining and Clip Construction}

We construct MeetingToM through a deterministic, task-driven pipeline. For each meeting, we first mine candidate temporal spans for the three task families and then instantiate non-overlapping clips within each meeting. For the subject-level \textsc{State} task, we mine interaction-based windows in which a target participant is visibly involved as a speaker or listener. For the dyadic \textsc{You} task, we locate utterances containing second-person pronouns and retain cases whose addressees can be resolved from surrounding verbal and visual context. For the group-level \textsc{Consensus} task, we use AMI decision timestamps as anchors and further identify proposal-response structures and agreement markers around decision moments.

Task~1 uses 5-second close-up clips centered on the target participant, while Tasks~2 and~3 use 50--70 second clips, preferring 50 seconds when sufficient context is available. For each selected segment, we synchronously cut the Corner view, all Closeup views, audio, and transcript using the same start and end timestamps with \texttt{ffmpeg}. Mosaic views are constructed from synchronized close-up views when required by the task. This procedure ensures temporal alignment between verbal and non-verbal evidence.

\subsection{GPT-assisted Question Generation}

GPT-5 is used only to draft candidate questions and task-specific options; it does not determine gold labels. Question generation follows fixed task formats, with only task-specific variables filled into predefined placeholders, such as the target participant, trigger utterance, addressee candidates, or consensus moment.

Generated questions are discarded if they are ambiguous, underspecified, inconsistent with the task definition, or not answerable from the provided clip. Human annotators assign the final answers based on the original video and transcript rather than on GPT-generated rationales.

\subsection{Annotation Protocol and Quality Control}

Each instance is independently labeled by two trained annotators with access to the corresponding video clip and transcript. Annotators are instructed to base decisions on observable evidence, including utterance content, turn-taking, gaze, facial expression, posture, gesture, response timing, and interactional context. They are also instructed to label only episode-bound states or relations and to avoid inferring stable personality traits, long-term preferences, or sensitive attributes.

When the two annotators disagree, a third annotator adjudicates the instance. Instances without a reliable final label are discarded. Annotators complete a pilot calibration stage in which guidelines and examples are refined before final annotation. We additionally remove samples with temporal misalignment, missing views, insufficient evidence, unresolved ambiguity, or questions whose answers depend on information outside the provided clip.

\subsection{Leakage Control and Prompt Consistency}

Original AMI annotations and metadata are used only for benchmark construction and validation, and gold labels are used only for scoring. At evaluation time, the default model input contains only the visual input, aligned transcript, and task question. AMI dialogue acts, participant roles, meeting phases, and other auxiliary annotations are excluded unless explicitly introduced in controlled ablation settings.

To minimize prompt-specific bias, all evaluation prompts are predefined from the task definitions and applied uniformly across models. We do not perform model-specific prompt tuning or post-hoc prompt optimization based on experimental results.

\subsection{Model Input Formatting and Video Ingestion}
\label{app:video_ingestion}

Our evaluation uses two ingestion pathways depending on model capability. For MLLMs that natively support video inputs, we provide the video clips directly together with the aligned transcript and the task prompt. All information required for evaluation is consolidated in task-specific \texttt{.jsonl} files.

For GPT-5, which does not support direct video inputs in our setup, we convert each clip into a sequence of uniformly sampled frames at 0.5 FPS and feed these frames together with the same transcript and prompt.

\section{Additional Prompting Diagnostics}
\label{app:diagnostics}

\subsection{Additional Open-Source Model Results}
\label{app:open_prompting}

The prompting analysis in the main text shows that thinking prompts do not provide uniform gains across tasks or models. To further examine this pattern beyond the models reported in Table~\ref{tab:prompting}, we evaluate three additional open-source MLLMs: InternVL2.5-26B, InternVL3.5-8B, and LLaVA-Video-7B-Qwen2. As shown in Table~\ref{tab:multi_model_prompting}, prompting effects remain non-uniform across these models, further supporting the observation that such strategies are task- and model-dependent.

\begin{table*}[t]
\centering
\scriptsize
\caption{
Additional prompting results on open-source MLLMs.
Colored arrows indicate absolute changes relative to No Prompt.
}
\setlength{\tabcolsep}{4.5pt}
\renewcommand{\arraystretch}{1.12}
\begin{tabular*}{\textwidth}{@{\extracolsep{\fill}}llccccc}
\toprule
Model & Prompt & Task~1 & Task~2.1 & Task~2.2 & Task~3.1 & Task~3.2 \\
\midrule
\multirow{3}{*}{InternVL2.5-26B}
& No Prompt 
& 30.67 
& 10.33 
& 48.33 
& 17.33 
& 25.00 \\
& Task-CoT 
& 42.17 \textcolor{pos}{\scriptsize{$\uparrow$11.50}} 
& 17.00 \textcolor{pos}{\scriptsize{$\uparrow$6.67}} 
& 47.33 \textcolor{neg}{\scriptsize{$\downarrow$1.00}} 
& 15.67 \textcolor{neg}{\scriptsize{$\downarrow$1.67}} 
& 42.55 \textcolor{pos}{\scriptsize{$\uparrow$17.55}} \\
& ECoT 
& 36.83 \textcolor{pos}{\scriptsize{$\uparrow$6.17}} 
& 23.33 \textcolor{pos}{\scriptsize{$\uparrow$13.00}} 
& 48.33 {\scriptsize($\pm$0.00)} 
& 16.00 \textcolor{neg}{\scriptsize{$\downarrow$1.33}} 
& 41.67 \textcolor{pos}{\scriptsize{$\uparrow$16.67}} \\
\midrule
\multirow{3}{*}{InternVL3.5-8B}
& No Prompt 
& 31.00 
& 22.33 
& 48.00 
& 18.00 
& 70.37 \\
& Task-CoT 
& 28.33 \textcolor{neg}{\scriptsize{$\downarrow$2.67}} 
& 15.33 \textcolor{neg}{\scriptsize{$\downarrow$7.00}} 
& 44.00 \textcolor{neg}{\scriptsize{$\downarrow$4.00}} 
& 15.67 \textcolor{neg}{\scriptsize{$\downarrow$2.33}} 
& 27.66 \textcolor{neg}{\scriptsize{$\downarrow$42.71}} \\
& ECoT 
& 31.83 \textcolor{pos}{\scriptsize{$\uparrow$0.83}} 
& 19.00 \textcolor{neg}{\scriptsize{$\downarrow$3.33}} 
& 37.67 \textcolor{neg}{\scriptsize{$\downarrow$10.33}} 
& 18.33 \textcolor{pos}{\scriptsize{$\uparrow$0.33}} 
& 47.27 \textcolor{neg}{\scriptsize{$\downarrow$23.10}} \\
\midrule
\multirow{3}{*}{LLaVA-Video-7B-Qwen2}
& No Prompt 
& 26.83 
& 19.00 
& 44.67 
& 58.67 
& 0.00 \\
& Task-CoT 
& 41.33 \textcolor{pos}{\scriptsize{$\uparrow$14.50}} 
& 22.00 \textcolor{pos}{\scriptsize{$\uparrow$3.00}} 
& 44.00 \textcolor{neg}{\scriptsize{$\downarrow$0.67}} 
& 53.67 \textcolor{neg}{\scriptsize{$\downarrow$5.00}} 
& 1.86 \textcolor{pos}{\scriptsize{$\uparrow$1.86}} \\
& ECoT 
& 29.67 \textcolor{pos}{\scriptsize{$\uparrow$2.83}} 
& 24.00 \textcolor{pos}{\scriptsize{$\uparrow$5.00}} 
& 44.33 \textcolor{neg}{\scriptsize{$\downarrow$0.33}} 
& 49.33 \textcolor{neg}{\scriptsize{$\downarrow$9.33}} 
& 3.38 \textcolor{pos}{\scriptsize{$\uparrow$3.38}} \\
\bottomrule
\end{tabular*}
\label{tab:multi_model_prompting}
\end{table*}

%

\section{Potential Assistive Applications}
\label{app:applications}

Meeting-grounded social reasoning may support future assistive meeting technologies, such as helping users locate moments of uncertainty, unresolved disagreement, possible misunderstanding, or decision points that require follow-up. Such systems could reduce post-meeting recap costs and provide contextual summaries or background explanations to support shared understanding. However, these applications should remain human-in-the-loop and uncertainty-aware: model outputs should be treated as prompts for reflection and clarification, not as definitive judgments about participants' private mental states, hidden intentions, or workplace performance.

\section{Ethical Considerations and Broader Impacts}
\label{app:ethics}

MeetingToM is intended as an evaluation benchmark rather than a deployment framework for inferring people's hidden intentions or attitudes in real-world settings. The labels in MeetingToM should be interpreted as systematic third-person inferences from observable evidence under a fixed annotation protocol, not as definitive ground truth about participants' inner mental states.

This distinction is important because Theory of Mind concerns latent states such as beliefs, intentions, uncertainty, and attitudes, which are not directly observable and must be inferred from behavior, language, and context~\cite{premack1978chimpanzee,chuey2025theory}. Human social inference is also fallible: people often overestimate how visible their internal states are to others, known as the illusion of transparency~\cite{gilovich1998illusion}, and even deception judgments are only modestly above chance in many real-time settings~\cite{bond2006accuracy}. Thus, ambiguity is not merely an annotation artifact but an inherent property of social inference.

This concern is especially relevant for AI systems. Recent work shows that strong models can perform well on constrained ToM benchmarks, yet their capabilities remain uneven and can lag behind human performance~\cite{strachan2024testing,jones-etal-2024-comparing-humans,chen2024tombench}. Other analyses further argue that standard benchmarks may capture only limited forms of ToM-like behavior rather than robust, functional social understanding~\cite{riemer2024position}. Accordingly, high benchmark accuracy should not be interpreted as reliable access to another person's private mental state.

MeetingToM may also reflect representational and social bias. Because it is built on structured professional meetings, it may privilege particular norms of turn-taking, disagreement, attentiveness, and affect display. Systems evaluated on this benchmark may therefore generalize poorly to informal conversations, cross-cultural interactions, or neurodiverse communication styles, where signals of hesitation, dissent, engagement, or alignment may be expressed differently.

A related risk is normative labeling. Our task formulation operationalizes complex social behavior into a finite set of mental-state and consensus categories. This abstraction enables controlled evaluation, but it should not be mistaken for a claim that there is always a single correct interpretation of another person's internal state. MeetingToM is best understood as a structured probe of model behavior under a fixed annotation scheme, not as a definitive ontology of human mental states.

We also note the possibility of dual use and misuse. Similar methods could be repurposed for workplace surveillance, automated assessment of employee engagement, or speculative judgment of disagreement, confidence, or attitude in professional settings. Such uses could disadvantage individuals whose communication styles differ from dominant norms or encourage overconfident inferences from incomplete evidence. We do not view MeetingToM as supporting high-stakes decision-making about real people, and benchmark performance should not be interpreted as evidence that such systems are ready for deployment in employment, educational, or managerial contexts.

Finally, prompting methods such as CoT or ECoT may improve benchmark performance while also encouraging fluent rationalizations for socially sensitive judgments. Such explanations can appear coherent even when weakly grounded in evidence. For this reason, prompting gains should not be read as evidence of calibrated, transparent, or unbiased social understanding.

Overall, MeetingToM should primarily be used to study model limitations, multimodal grounding, and evaluation methodology. Results should be interpreted together with task scope, class imbalance, annotation subjectivity, and domain specificity.

\section{Ecological Scope, Temporal Context, and Modern Meeting Settings}
\label{app:scope_limitations}

MeetingToM is built on the AMI Meeting Corpus. AMI contains in-person scenario meetings with synchronized multi-view video, transcripts, and relatively clear interaction structures. This setting provides a useful basis for constructing social reasoning tasks in meetings: participants share the same physical space, their roles and meeting goals are relatively explicit, and cues such as speech, posture, gaze, body orientation, and turn-taking can be observed in a relatively complete way. Therefore, MeetingToM is best understood as a controlled benchmark for evaluating latent social-state reasoning under meeting conditions where multimodal cues are rich and the interaction structure is relatively clear.

MeetingToM focuses on structured meeting scenarios because they provide rich multimodal signals, clear interaction structure, and practical relevance. This choice improves controllability and annotation consistency, but may limit generalization to other social settings, such as informal conversations or online and hybrid meetings. Our current design uses task-centered segments rather than full meetings to balance annotation reliability, precise ground-truth alignment, and current MLLM video-context limits. This does not reduce the benchmark to local cue detection: Tasks~2 and~3 still require multi-turn context, including preceding proposals, responses, gaze shifts, and later clarifications. Extending the benchmark to longer meeting-level dynamics remains an important direction for future work.

Recent meeting benchmarks further clarify this positioning. QMSum, MeetingBank, and VCSum mainly focus on meeting summarization~\citep{zhong2021qmsum, hu2023meetingbank, wu2023vcsum}; MUG covers general meeting understanding and generation tasks such as topic segmentation, summarization, keyphrase extraction, and action item detection~\citep{zhang2023mug}; M2MeT focuses on multi-channel multi-party meeting transcription, including speaker diarization and multi-speaker ASR~\citep{yu2022m2met}; MISP-Meeting / MISP 2025 support multimodal Mandarin meeting transcription, summarization, audio-visual speaker diarization, speech recognition, and joint diarization-recognition~\citep{chen2025misp, gao2025multimodal}; and ELITR-Bench further moves toward meeting assistant evaluation over long noisy ASR transcripts~\citep{thonet2025elitr}. These works have advanced models' ability to process meeting content, identify speakers, generate summaries, extract action items, and support meeting assistants. Overall, recent meeting benchmarks mainly treat meetings as information records or workflow objects: what was said, who said it, what should be summarized, and what tasks should be carried out.

However, real meetings are not only information exchange processes. They are also processes of social collaboration and group coordination. Participants do not always express their true stance directly. Due to politeness, power relations, group pressure, conflict avoidance, or time pressure, they may manage disagreement and consensus through silence, brief agreement, hesitation, softened expressions, or delayed responses. For AI systems intended to operate in real social settings, recognizing meeting content alone is not enough. A socially aware system also needs to understand how participants respond to one another, who only appears to agree, who may still hold reservations, whether a decision has genuinely gained group support, and whether smooth visible progress masks hidden disagreement. In this sense, social reasoning is not a secondary issue in meeting understanding, but a core capability for AI systems that aim to participate in collaborative, organizational, and social interaction settings.

MeetingToM is designed to address this gap. It treats meetings as socially situated interactions rather than simply as information carriers. It focuses not only on what was said, but also on how participants display uncertainty, support, reservation, disagreement, weak buy-in, or pseudo-consensus during interaction. In this sense, MeetingToM complements recent meeting benchmarks by systematically evaluating multimodal social-pragmatic reasoning over latent interactional states.

This positioning also helps clarify the relation between AMI and modern remote or hybrid meetings. Meetings in AMI differ from contemporary platform-mediated meetings. In remote or hybrid meetings, camera placement, participant visibility, gaze interpretation, turn-taking, and participation patterns may all change. Yet what changes is not whether these social phenomena exist, but the cue system through which they are observed, interpreted, and annotated. Individual uncertainty, directed stance, weak buy-in, hidden dissent, and apparent consensus can still arise in remote and hybrid meetings. In in-person meetings, these states may be expressed through facial expression, gaze, posture, gesture, body orientation, turn-taking, and shared physical space. In video-mediated meetings, they may shift toward platform-mediated cues, such as camera status, mute/unmute behavior, response latency, chat feedback, emoji reactions, screen sharing, comments on shared documents, and differences in the meeting layout visible to different participants.

This shift affects the three levels of MeetingToM in different ways. For subject-level reasoning, hesitation, confusion, or disengagement may no longer be expressed mainly through posture or gesture, but through pauses, hedging, self-repairs, delayed responses, brief acknowledgments, camera-off silence, or chat-based participation. For dyadic-level reasoning, gaze and body orientation become less stable evidence for addressee relations. Models may need to rely more on explicit names, quoted references, reply structure, adjacent turns, chat replies, or actions during screen sharing. For group-level reasoning, pseudo-consensus may still occur, but it may appear through silence, minimal verbal agreement, ignored chat objections, unresolved comments on shared documents, or asymmetric participation between in-room and remote attendees. Hybrid meetings are especially complex because some alignment cues may be locally visible only to participants in the same room, while remaining invisible or delayed for remote participants.

These differences also imply that extending MeetingToM to remote and hybrid meetings would require more than simply replacing AMI with a newer corpus. It would require new annotation schemes for platform-mediated cues, uneven participant visibility, artifact-centered interaction, and asymmetric participation structures. For example, camera-off participation may indicate disengagement, privacy preference, bandwidth constraints, or platform norms depending on context; chat objections may be highly salient to some participants but invisible or ignored by others; and shared-document comments may function as delayed forms of disagreement or repair. These cues are not weaker versions of in-room signals, but different interactional resources that reshape how social states become observable.

Therefore, the value of MeetingToM lies in making the social reasoning dimension of meeting understanding explicit and systematic. Existing meeting benchmarks have advanced models' understanding of meeting content, speaker attribution, summarization, and action item extraction. Yet collaboration in real meetings is not composed only of explicit information; it also involves how participants respond to one another, adjust their stances, withhold disagreement, and form group consensus. AMI provides a research starting point where multimodal cues are relatively rich and the interaction structure is relatively clear. Extending this framework to remote and hybrid meetings is not only a matter of introducing newer meeting data, but also of understanding how platform-mediated meetings reshape the visibility, sharedness, and interpretability of social cues. In other words, modern meeting settings do not make social reasoning less relevant; rather, they make it appear through new cue forms. Models need to connect language, vision, platform behavior, and interaction structure in order to better understand relations, stances, and group states in meetings.

\section{Mapping Meeting Mental States to AMI Annotations}
\label{app:ami-mapping}

\begin{table*}[t]
\centering
\footnotesize
\caption{Meeting mental state taxonomy in MeetingToM. Engagement, valence, and expression describe coarse affective–behavioral profiles; the rightmost column shows how each state is operationalized using existing AMI-style annotation layers.}
\setlength{\tabcolsep}{3.3pt}
\begin{tabular}{
p{0.16\textwidth}
p{0.08\textwidth}
p{0.08\textwidth}
p{0.08\textwidth}
p{0.52\textwidth}
}
\toprule
Meeting mental state & Engagement & Valence & Expression & Key AMI-derived cues \\
\midrule
Active Engagement 
& High 
& Positive 
& High 
& Dense \textit{dialogueActs} involving informing, suggesting, and elaborating; frequent nods in \textit{headGesture}; task-oriented \textit{handGesture} such as pointing or writing; \textit{focus} on the current speaker or shared artefacts; forward-leaning patterns in \textit{movement} within the corresponding \textit{segments}. \\

Focused Listening 
& High 
& Positive 
& Low 
& Long stretches with few floor-taking \textit{dialogueActs} beyond backchannels; sustained \textit{focus} on the current speaker; subtle nods in \textit{headGesture}; minimal \textit{handGesture} and low-amplitude \textit{movement}, while the content of \textit{words} stays on-topic. \\

Supportive Endorsement 
& Medium 
& Positive 
& Medium 
& Short agreement or positive-assessment \textit{dialogueActs}; nodding \textit{headGesture} aligned with others’ turns; occasional affiliative \textit{handGesture}; stable \textit{focus} on the proposer during supportive comments; local \textit{segments} that follow another participant’s proposal in the same \textit{topic}. \\

Confused Bewilderment
& Medium 
& Negative 
& Low 
& Clarification or repair-initiation \textit{dialogueActs}; increased \textit{disfluency} in the \textit{words}; tilts or brief freezes in \textit{headGesture}; unstable \textit{focus} alternating between speaker and materials; often appearing when \textit{topics} shift or new information is introduced. \\

Hesitation
& Medium 
& Negative 
& Low 
& Filled pauses and self-repairs in \textit{disfluency}; delayed response after incoming \textit{dialogueActs}; micro-shifts in \textit{headGesture} and \textit{focus}; small self-adaptive \textit{handGesture} or posture changes within a single \textit{segment}. \\

Cognitive Conflict 
& High 
& Negative 
& High 
& Disagreement or rejection \textit{dialogueActs}; overlaps in turn-taking; head shakes in \textit{headGesture}; expansive \textit{handGesture}; rapid \textit{focus} shifts between opponents and shared artefacts around contested \textit{topics}. \\

Disengaged Withdrawal 
& Low 
& Negative 
& Low 
& Sparse \textit{dialogueActs}; \textit{focus} on table or objects rather than speakers; minimal task-related \textit{handGesture}; leaning-back or slumped \textit{movement}; off-topic or absent \textit{words}. \\
\bottomrule
\end{tabular}
\label{tab:mental-states-compact}
\end{table*}

\subsection{Prompt Templates}
\label{app:prompt_templates}

\begin{figure*}[t]
  \centering
  \includegraphics[width=1\linewidth]{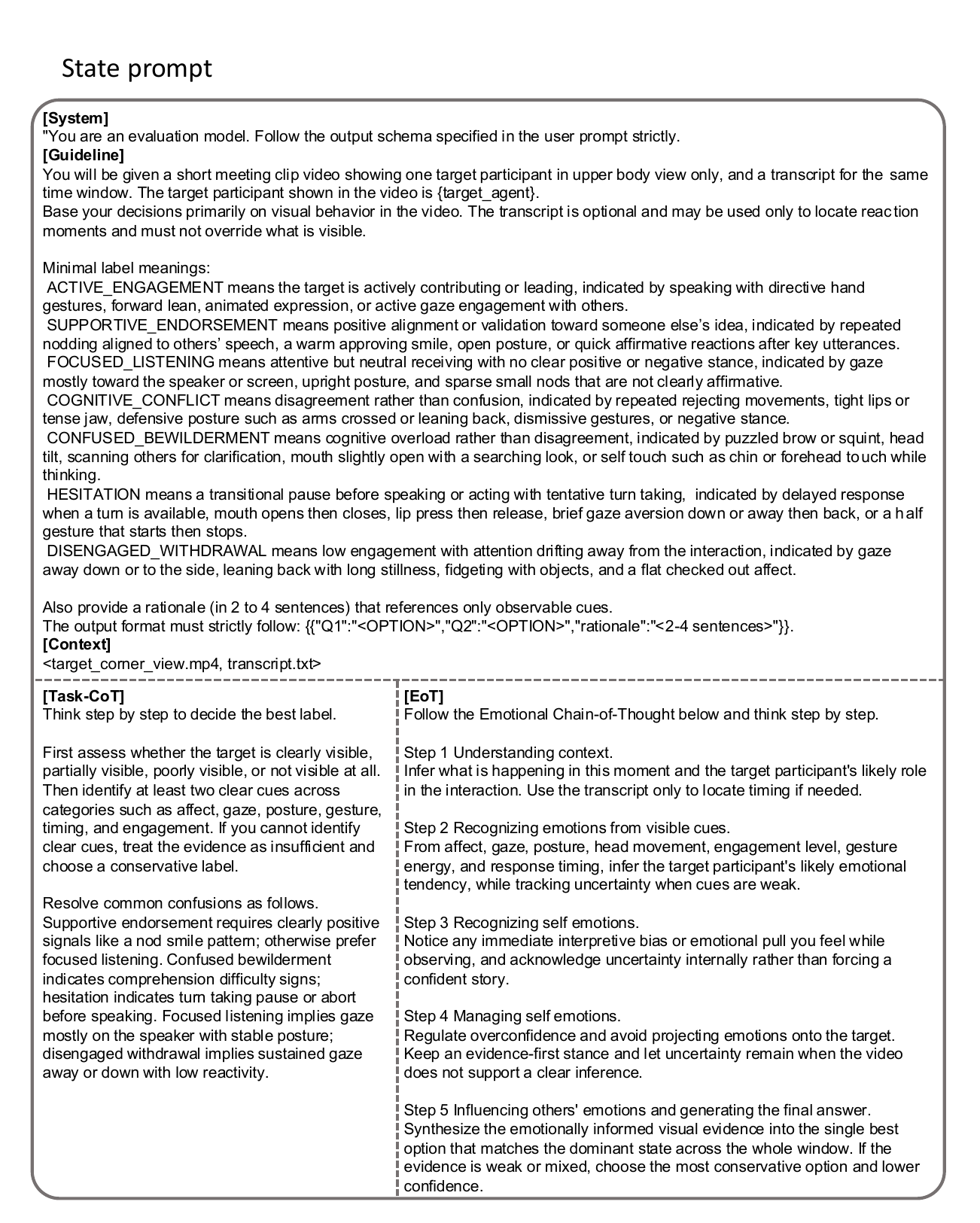} 
  \label{fig:prompt_template_1}
\end{figure*}

\begin{figure*}[t]
  \centering
  \includegraphics[width=1\linewidth]{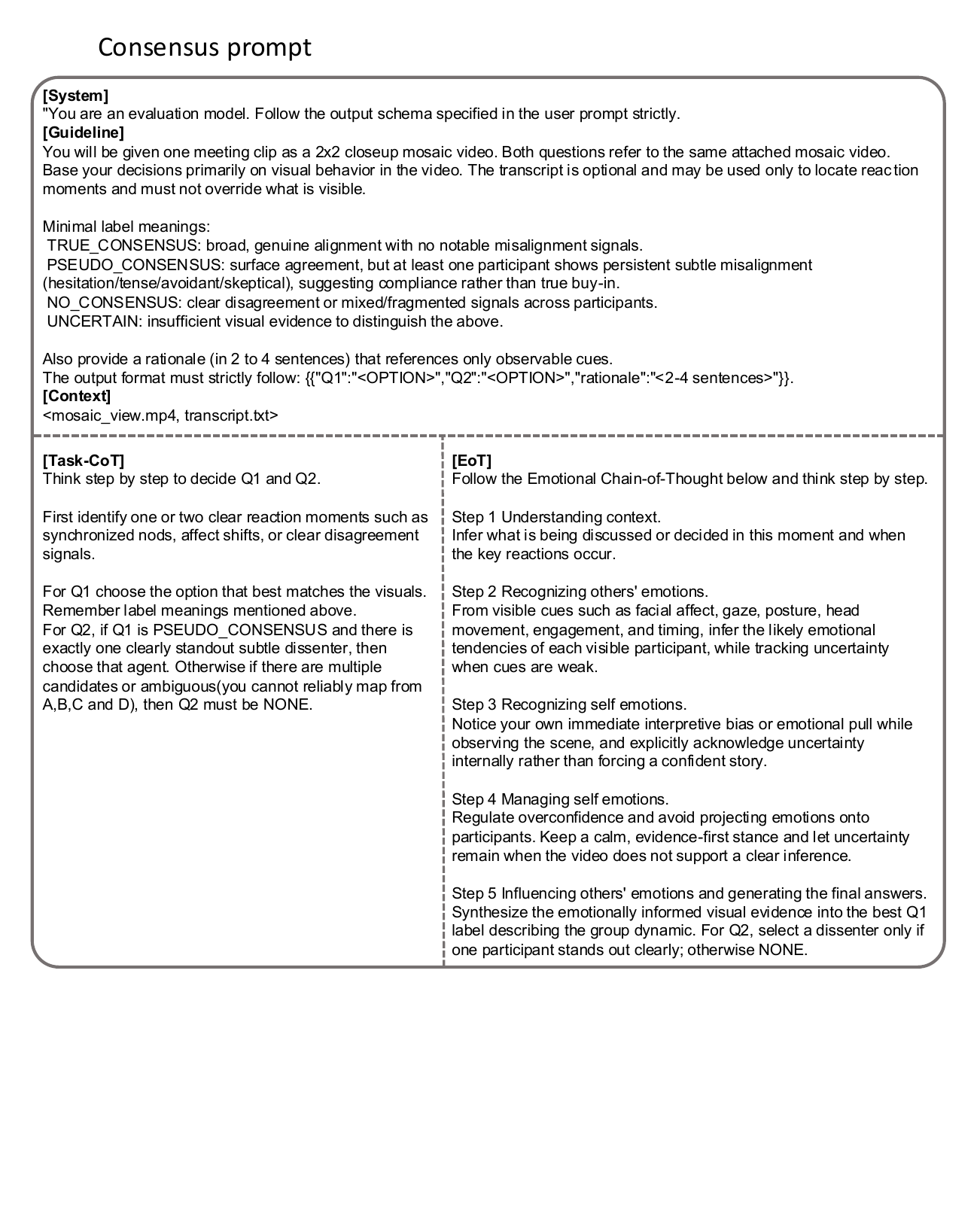} 
  \label{fig:prompt_template_2}
\end{figure*}

\begin{figure*}[t]
  \centering
  \includegraphics[width=1\linewidth]{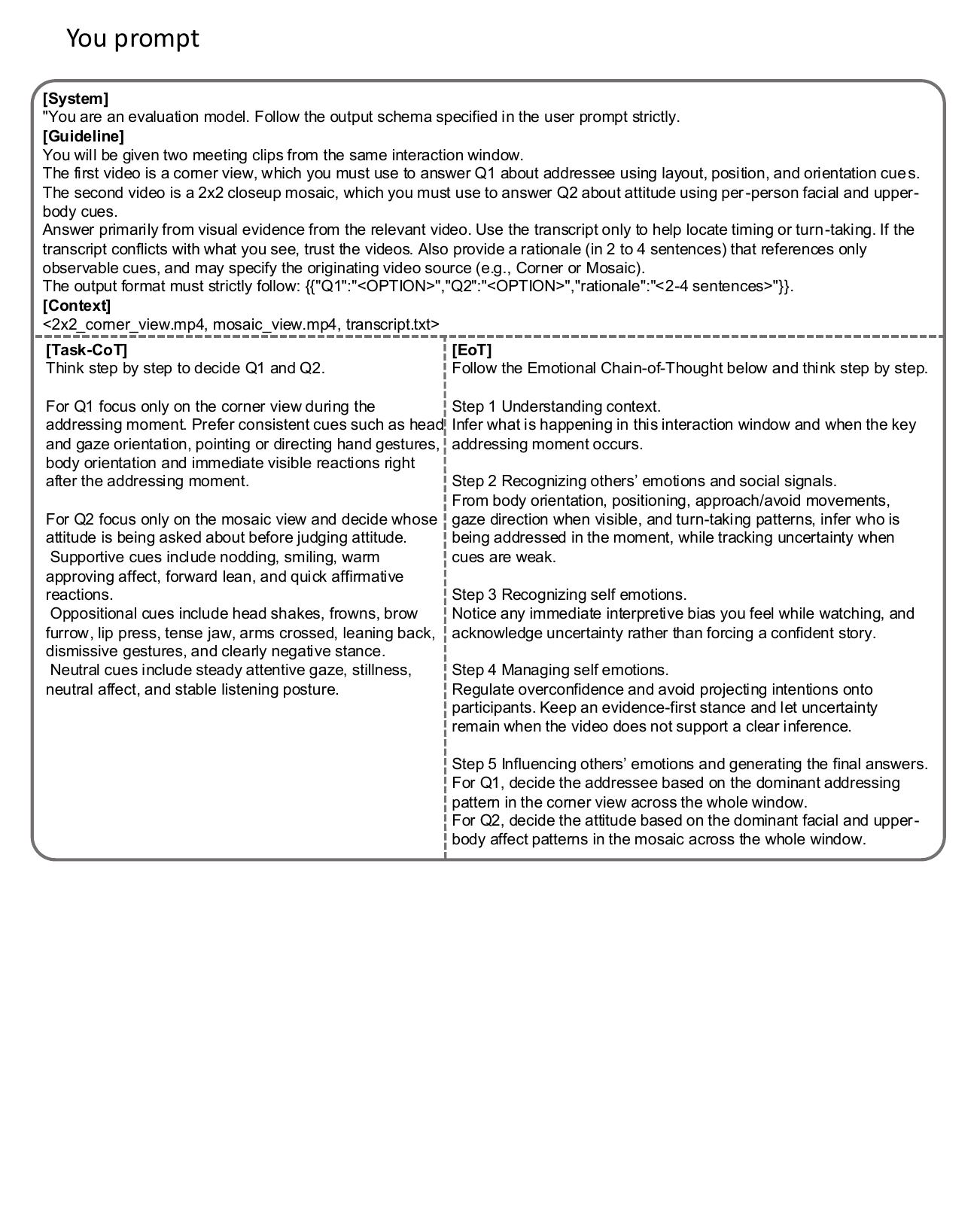} 
  \caption {MeetingToM CoT ECoT Prompt Examples}
  \label{fig:prompt_template_3}
\end{figure*}

\subsection{Annotation Interface and Guidelines}
\label{app:annotation_guidelines}

\begin{figure*}[t]
  \centering
  \includegraphics[width=1\linewidth]{figures/Annotationpage.pdf} 
  \caption {MeetingToM Annotation Page}
  \label{fig:annotation_page}
\end{figure*}

\begin{figure*}[t]
  \centering
  \includegraphics[width=1\linewidth]{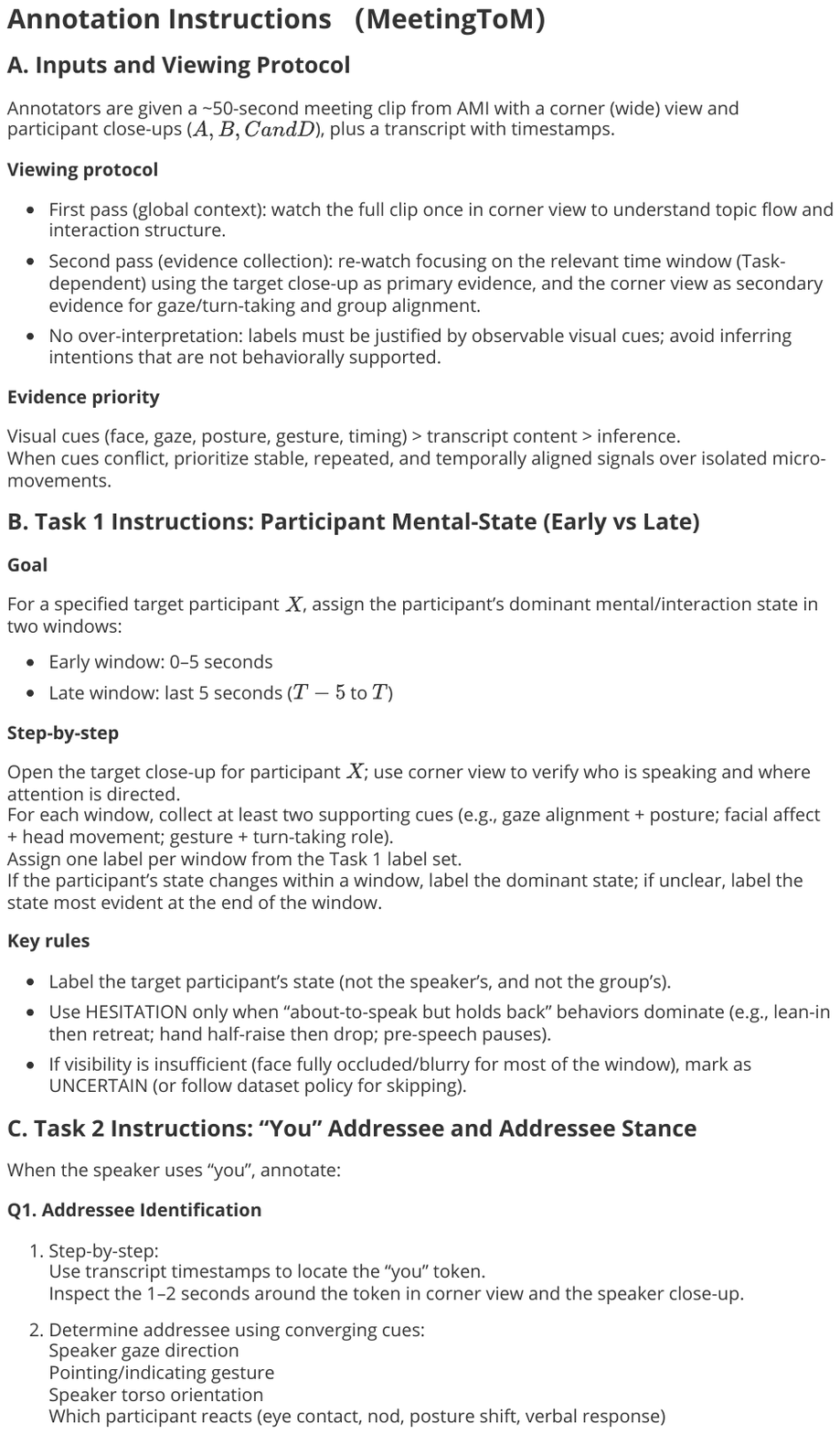} 
  \label{fig:annotation_guideline_1}
\end{figure*}

\begin{figure*}[t]
  \centering
  \includegraphics[width=1\linewidth]{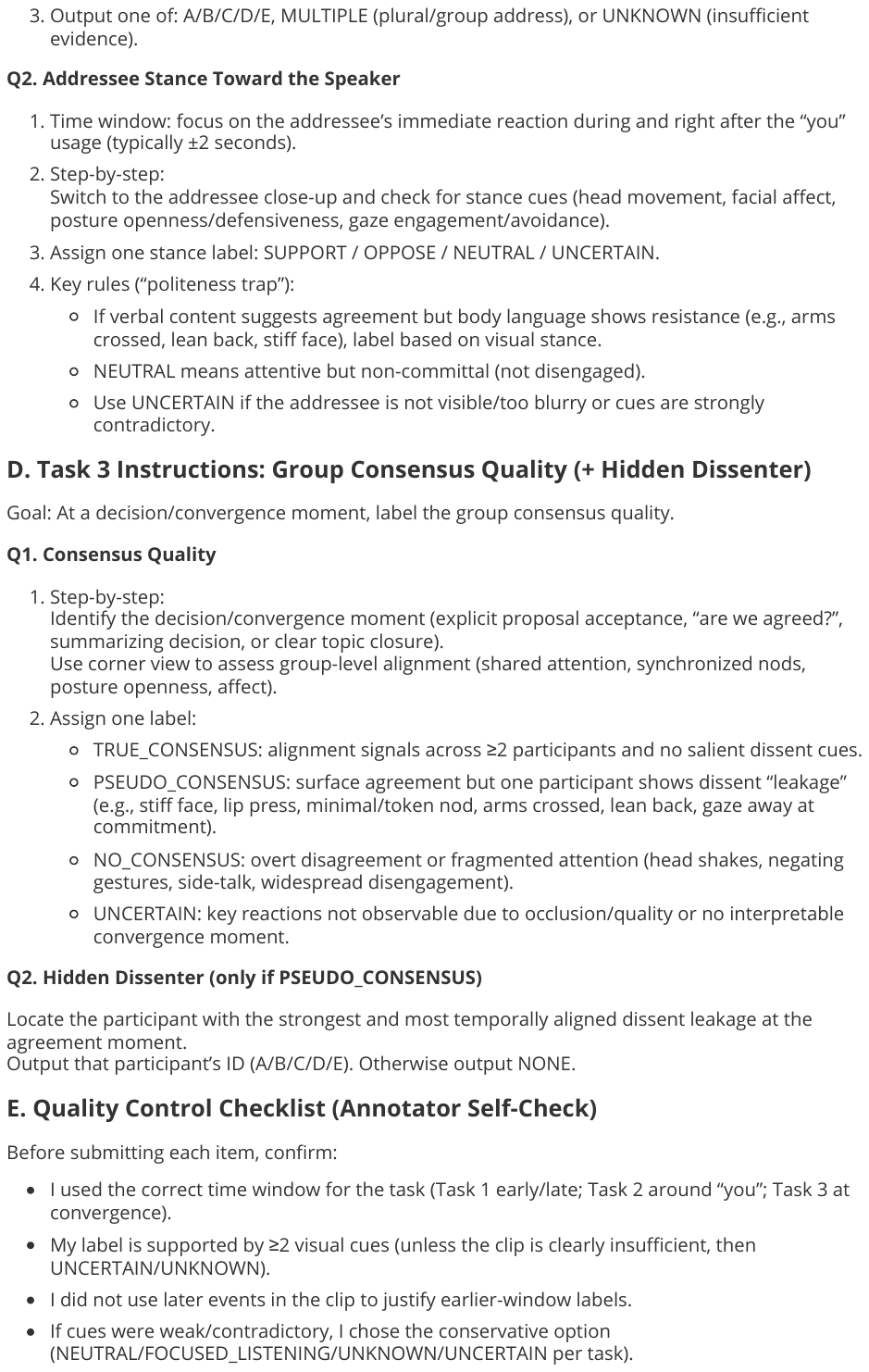} 
  \caption {MeetingToM Annotation Guidelines}
  \label{fig:annotation_guideline_2}
\end{figure*}


\end{document}